# Adversarial Feature Sampling Learning for Efficient Visual Tracking

Yingjie Yin[1,2], Lei Zhang[1], De Xu[2] and Xingang Wang[2]

1.The Hong Kong Polytechnic University, Hung Hom, Kowloon, Hong Kong

2.Institute of Automation, Chinese Academy of Sciences, BEIJING, CHINA

haidaying.jie@163.com; cslzhang@comp.polyu.edu.hk; de.xu@ia.ac.cn; xingang.wang@ia.ac.cn

**Abstract**

The tracking-by-detection framework usually consist of two stages: drawing samples around the target object in the first stage and classifying each sample as the target object or background in the second stage. Current popular trackers based on tracking-by-detection framework typically draw samples in the raw image as the inputs of deep convolution networks in the first stage, which usually results in high computational burden and low running speed. In this paper, we propose a new visual tracking method based on adversarial feature sampling learning to address this problem. Only one cropped image around the target object is input into the designed deep convolution network and the samples is sampled on the feature maps of the network by spatial bilinear resampling. Positive samples obtained by sampling are severely overlapped in space because of deep convolutional features' lower resolution. In order to capture target's rich appearance variations, a generative adversarial network is integrated into our network framework to augment positive samples and improve the tracking performance. Extensive experiments on benchmark datasets demonstrate that the proposed method achieves a comparable performance to state-of-the-art trackers and accelerates tracking-by-detection trackers based on raw-image samples effectively.

## 1. Introduction

Visual tracking is one of important problems in computer vision and is widely used in intelligent video surveillance, human–computer interaction, visual navigation, and so on. Conventional tracking methods [1, 2, 3, 4, 5, 6] use low-level handcrafted features to follow target objects. Although they achieve computational efficiency and comparable tracking performance, they are still limited in solving the tracking obstacles such as motion blur, occlusion, illumination change, and background clutter because of their insufficient feature representation.

Recently, tracking methods [7, 8, 9, 10, 11, 12] using deep convolutional neural networks (CNNs) have been proposed for robust tracking and the tracking performance is vastly improved with the help of rich feature representation. Some current state-of-the-art trackers are typically based on a two-stage deep-learning-based tracking-by-detection framework. A sparse set of samples around the target object are drawn in the fist stage and the second stage classifies each sample as either the target object or as the background using a deep convolution network. Hong et al. [12] proposed a tracking-by-detection tracking algorithm which learns discriminative saliency map using CNN features and an online SVM and achieved substantial performance gain. Nam *et al*. [10] proposed a tracking-by-detection tracking algorithm with designed Multi-Domain Network trained with tracking video datasets such as [13, 14] and achieved the better performance compared to the traditional trackers. Song et al. [7] further increased the tracking performance by introducing adversarial learning into Nam's Multi-Domain Network to enrich the target appearances in the feature space and augment the positive samples.

Though these deep-learning-based tracking-by-detection trackers achieved favorable performance on recent tracking benchmarks [13, 14, 15, 16], their efficiency is limited in the first stage of the tracking-by-detection framework. Samples in the raw image are typically drawn as the inputs of deep convolution networks and each raw-image sample needs to be evaluated by the deep neural network separately, which results in high computational burden and low running speed. It is of great importance to investigate how to improve the efficiency of deep-learning-based tracking-by-detection framework by changing the sampling strategy in the first stage.

In this work, to deal with the issues raised above, we propose a novel tracker a new visual tracking method based adversarial feature sampling learning (AFSL). For a deep classification network, such as the VGG-16 model [17], we just input one cropped image around the target object to generate deep convolutional features. Then the features of samples are drawn by the spatial bilinear resampling (SBR) in the generated deep convolutional features. In the deep CNN, the high-level features contain well semantic information but are not sensitive to the translations. The low-level features have high resolutions and better position information but could not represent the features well. In order to obtain robust sampling features for each sample, the sampling features from high-level feature and low-level feature are fused to obtain the final sampling deep convolutional features. Compared with raw-image samples, sampling deep convolutional features is more efficiency because the deep CNN evaluate only one cropped image rather than quite a number of raw-image samples. The spatial resolution of the deep convolutional features is far lower than the input image, so the sampling deep convolutional features of the positive samples are highly spatially overlapped, and they fail to capture rich appearance variations. To augment positive samples, a generative adversarial network is integrated into our network framework to randomly generate masks, which are applied to adaptively dropout the sampling deep convolutional features to capture a variety of appearance changes over a long temporal span.

The main contributions of this paper are as follows.
(1) we propose a new visual tracking method using sampling deep convolutional features to improve the efficiency of deep-learning-based tracking-by-detection trackers effectively.
(2) A generative adversarial network is integrated into our network framework to augment the sampling deep convolutional features of positive samples to capture a variety of object's appearance changes over a long temporal span.
(3) Extensive experiments on benchmark datasets demonstrate that the proposed method achieves a comparable performance to state-of-the-art trackers and accelerates tracking-by-detection trackers based on raw-image samples effectively.

**2. Related work**

Visual tracking has been studied extensively over the past decades. Comprehensive surveys of this active research field can be found in [14, 15, 18]. In this section, we mainly discuss the methods closely related to this work in terms of online deep-learning-based trackers and generative adversarial learning.

**2.1 online deep-learning-based trackers**

Deep learning has fueled great strides in a variety of computer vision problems, such as visual tracking [7, 8, 9, 10, 11, 12], object detection [19, 20, 21, 22], action recognition [23, 24] and semantic segmentation [22, 25, 26]. Visual tracking is one of the fundamental problems in computer vision. Recent trackers based on online deep learning have outperformed previous low-level feature-based trackers. These state-of-the-art trackers are mainly based on the one-stage regression framework or the two-stage classification framework. As one of the most representative types of the one-stage regression framework, the trackers based on correlation filters [27, 28, 29, 30] are trained online to regress all the circular-shifted version of the input features into soft labels of a Gaussian function rather than binary labels for discriminative classifier learning. Ma et al. [27] proposed a tracking method based on hierarchical convolutional features and correction filters. The correlation filters are adaptively learned online on convolutional layers from different levels to encode the target appearance and the maximum response of each layer is inferred hierarchically to locate targets. Qi et al. [28] proposed a CNN based tracking framework which uses an adaptive online decision learning algorithm to hedge weak trackers, obtained by correlation filters on CNN feature maps, into a stronger one to achieve better results. The correlation filter methods obtain a number of training samples by cyclic shift and can fast solve the filters online by fast Fourier transform algorithm, however, their performances are usually affected by boundary effects.

In contrast, the deep-learning-based two-stage classification framework treats the tracking task as a binary classification problem. In the first stage, many samples are drawn around the target object and these samples are classified as the target object or background in the second stage. The two-stage trackers emphasize on a discriminative boundary between the samples of the target object and background. Wang et al. [31] proposed a deep learning tracker (DLT) which uses a stacked denoising autoencoder (SDAE) [32] as a feature extractor during the online tracking process to train a classification neural network to distinguish the tracked object from the background. Gao et al. [33] proposed a deep relative tracking algorithm through the convolutional neural network and the proposed method can effectively exploit the relative relationship among image patches from both foreground and background for object appearance modeling. Nam et al. [10] proposed a tracking algorithm called MDNet based on a CNN trained in a multi-domain learning framework. MDNet learns domain independent representations from pretraining and captures domain-specific information through online learning during tracking. Song et al. [7] proposed visual tracking via adversarial learning (VITAL) which integrated adversarial learning into the tracking-by-detection tracking framework to enrich the target appearances in the feature space and augment the positive samples. Though these trackers based on deep-learning-based two-stage classification framework can achieve better accuracy, they usually surfer high computational burden and run at low speed because a lot of samples in the raw image are typically drawn as the inputs of deep convolution networks and each raw-image sample needs to be evaluated separately.

**2.2 Generative adversarial learning**

Generative Adversarial Networks (GAN) are proposed by Goodfellow et al. in [34]. A GAN consists of a discriminative model and a generative model. The discriminative model aims to distinguish between real samples and generated samples, while the generative model is to generate fake samples as real as possible, making the discriminator believe that the fake samples are from real data. The generative model and the discriminator model are trained simultaneously with conflicting objectives.

So far, plenty of works have shown that GANs can play a significant role in various computer vision tasks, such as object detection [35], image super-resolution [36], semantic segmentation [37] and visual tracking [7, 38]. Li et al. [35] proposed a new Perceptual Generative Adversarial Network (Perceptual GAN) model that improves small object detection through narrowing representation difference of small objects from the large ones. Ledig et al. [38] proposed a generative adversarial network for image super-resolution (SRGAN) which is capable of inferring photo-realistic natural images for 4× upscaling factors. Souly et al. [37] developed a novel semi-supervised semantic segmentation approach employing Generative Adversarial Networks and the proposed model generates plausible synthetic images that supports the discriminator in the pixel-classification step. Song et al. [7] and Wang et al. [38] used adversarial learning to generate improved positive samples to promote the tracking performance.

**3. Proposed Algorithm**

Our proposed AFSL-based tracker consists of three modules. The first module is the base deep CNN which is used to generate deep convolutional features and the second module is a classification network based on sampling deep convolutional features. The last module is an adversarial feature generator network aiming at augment positive samples. The whole framework of our AFSL-based tracker is shown in Fig.1.

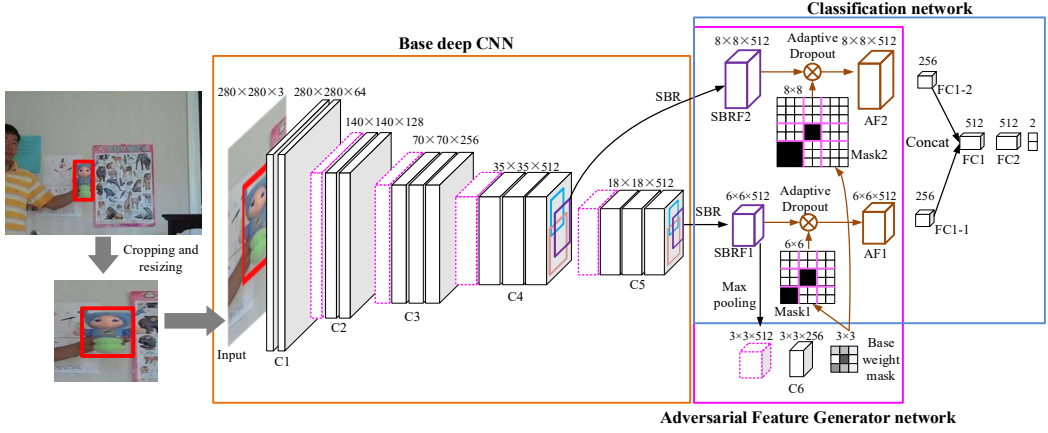

Figure 1. Overview of our proposed AFSL-based tracker.

### 3.1 Base deep CNN

The based deep CNN in our AFSL-based tracker is the first five groups of convolution layers of VGG-16 model [17]. Given an image $I_{t+1}$ at time $t+1$ and the target object's bounding box $B_t$ ($x_t$, $y_t$, $w_t$, $h_t$) at time $t$, we cropped an image patch $I_{t+1}^c$ around $B_t$ with a fixed cropped expansion rate $r_c$. Then the cropped image patch $I_{t+1}^c$ is resized to make the width and length of the target object keep a fixed value $L$. The size of the based deep CNN' input can be calculated by Equation (1).

$$W_{t+1}^{cr} = \frac{L}{w_t} \min(r_c w_t, W)$$
$$H_{t+1}^{cr} = \frac{L}{h_t} \min(r_c h_t, H)$$
(1)

where ($W_{t+1}^{cr}$, $H_{t+1}^{cr}$) is size of the cropped and resized image patch and ($W$, $H$) is the size of the original image. The cropped and resized image $I_{t+1}^{cr}$ is input into the based deep CNN to obtain the base convolutional features C4-3 and C5-3, which are corresponding to the last feature maps of groups C4 and C5 respectively in VGG-16 model. These base convolutional features will be used to generate the sampling deep convolutional features.

### 3.2 Sampling Deep Convolutional Features

Let $F \in \mathbb{R}^{H_F \times W_F \times C_F}$ be a based feature map, the SBR operation used to generate sampling feature map $F^S \in \mathbb{R}^{H_S \times W_S \times C_F}$ on $F$ can be expressed as follows:

$$F_{i'j'c}^S = \sum_{i=1}^{H_F} \sum_{j=1}^{W_F} F_{ijc} \max\{0, 1 - |y_{i'j'} - i|\} \max\{0, 1 - |x_{i'j'} - i|\} \quad (2)$$

where $F_{i'j'c}^S$ is the value of the $c$th channel at the spatial position ($i'$, $j'$) of $F^S$ and ($x_{i'j'}$, $y_{i'j'}$) is the sampling position of $F_{i'j'c}^S$ in the based feature map $F$. In the deep CNN, high-level features contain well semantic information and low-level features own better position information, so two sampling features from different levels are fused to describe one sample. Given a sampling bounding box $B'_{t+1}$ ($x'_{t+1}$, $y'_{t+1}$, $w'_{t+1}$, $h'_{t+1}$) in $I^{cr}_{t+1}$, we calculate its corresponding sampling bounding boxes on C4-3 and C5-3 by $B'_{t+1}/8$ and $B'_{t+1}/16$ respectively, where 8 and 16 are feature map strides in C4-3 and C5-3 respectively. As shown in Fig.2, sampling points in each level feature map are distributed in the sampling bounding box evenly and their values are calculated by Equation (2).

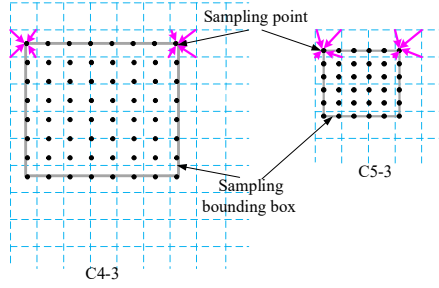

**Figure 2. The sampling process and the mask block partition.** The two rectangle boxes in C4-3 and C5-3 are corresponding to the same sampling bounding box in $I^{cr}_{t+1}$. The circle points are sampling points and their position coordinates are continuous. Each sampling point's value is calculated by its four neighbor points through SBR.

### 3.3 Adversarial learning for Sampling Positive Samples Augmentation

The traditional generative adversarial networks (GANs) [34] includes two parts, the discriminator network $D$ and the generative network $G$. The generator $G$ is learned to maximize the probability of $D$ classifying a generating image as a real image and the discriminator is leaned to avoid this mistake. The objective loss function for training $G$ and $D$ is defined as:

$$\mathcal{L} = \min_G \max_D \mathbb{E}_{x \sim P_{data}(x)}\left[\log D(x)\right] + \mathbb{E}_{z \sim P_{noise}(z)}\left[\log\left(1 - D(G(z))\right)\right] \tag{3}$$

where $z$ is a noise vector from a distribution $P_{noise}(z)$ and $x$ is a real image with a distribution $P_{data}(x)$. The training encourages $G$ to fit $P_{data}(x)$ so that $D$ will not be able to discriminate $x$ from $G(z)$.

It is not feasible to apply GANs in the tracking-by-detection framework directly because the input data to the framework are usually candidate object proposals rather than random noise and the classifier in tracking-by-detection framework is trained via supervised learning using labeled samples rather than unlabeled ones. To address this problem, Song et al. [7] proposed visual tracking via adversarial learning (VITAL) to narrow the gap between GANs and the tracking-by-detection framework. In VITAL, a generative network $G$ is used to randomly generate masks, which are applied to adaptively dropout input features to capture a variety of appearance changes. The objective function is defined as:

$$\mathcal{L}_{VITAL} = \min_G \max_D \mathbb{E}_{(C,M) \sim P(C,M)}\left[\log D(M \cdot C)\right] + \mathbb{E}_{C \sim P(C)}\left[\log\left(1 - D(G(C) \cdot C)\right)\right] \\ + \lambda \mathbb{E}_{(C,M) \sim P(C,M)}\left\|G(C) - M\right\|^2 \tag{4}$$

where $C$ is the input feature, $G(C)$ is the mask generated by the $G$ network and $M$ is the actual mask identifying the discriminative features.

Similar to VITAL, we design an adversarial feature generator network between sampling deep convolutional features and the classifier. The adversarial feature generator network will predict a base weight mask which is used to generate binary masks of different sizes for adaptively dropping out

sampling features. We define the objective function as:

$$\mathcal{L}_{SDCFAL} = \min_G \max_D \mathbb{E}_{(C^1,C^2,M) \sim P(C^1,C^2,M)} \left[ \log D\left( f_1(M, H_s^1, W_s^1) \cdot C^1, f_1(M, H_s^2, W_s^2) \cdot C^2 \right) \right]$$
$$+ \mathbb{E}_{C^1 \sim P(C^1)} \left[ \log\left(1 - D\left( f_2(G(C^1), H_s^1, W_s^1) \cdot C^1, f_2(G(C^1), H_s^2, W_s^2) \cdot C^2 \right) \right) \right] \quad (5)$$
$$+ \lambda \mathbb{E}_{(C^1,M) \sim P(C^1,M)} \left\| G(C^1) - M \right\|^2$$

where $D$ and $G$ correspond respectively to the classification network and adversarial feature generator network shown in Fig.1. $C^1$ and $C^2$ correspond respectively to sampling features SBRF1 and SBRF2 shown in Fig.1. $f_1(\cdot)$ and $f_2(\cdot)$ stand for the operations generating binary masks of suitable sizes for sampling features. $M$ is the reference mask. $G(C^1)$ is the base weight mask generated by the adversarial feature generator network. $H_s^1$ and $W_s^1$ are the height and width of SBRF1 and $H_s^2$ and $W_s^2$ are the height and width of SBRF2. The dot is the dropout operation on sampling features.

In the classification network $D$, two fully connected layers FC1-1 with 256 channels and FC1-2 with 256 channels are connected to the adversarial sampling features AF1 and AF2 shown in Fig.1 parallelly. Then FC1-1 and FC1-2 are concatenated as FC1 and another fully connected layer FC2 with 512 channels is connected to FC1. The last layer is a two-class output layer. The cross-entropy cost function shown in Equation (6) is used for training classification network $D$.

$$\mathcal{L}_D = -\frac{1}{N} \sum_{n=1}^{N} \sum_{i=1}^{2} y_{in} \log \left( \frac{e^{a_{in}}}{\sum_{k=1}^{2} e^{a_{kn}}} \right) \quad (6)$$

where $a_{in}$ is the output of the classification network $D$, $y_{in}$ is the label of the $n$th input sample and $N$ is the batch size.

In the adversarial feature generator network $G$, max pooling with kernel size 2×2 and stride 2 is operated on the sampling feature SBRF1, then a convolution layer C6 with kernel size $256 \times 1 \times 1 \times 512$ and stride 1 is followed. The base weight mask is the output of the last layer which is the convolution layer with kernel size $1 \times 1 \times 1 \times 256$ and stride 1. The values of the base weight mask are continuous and its spatial size is different from sampling features SBRF1 and SBRF2, so we define the operation $f_2(\cdot)$ to achieve binary masks of suitable sizes. In the operation $f_2(\cdot)$, several positions owning lower values in the base weight mask is found firstly and then the binary masks' values in the positions corresponding to these found positions in the base weight mask are set to 0 and values in other positions are set to 1. The positions corresponding to the $r_M$th row and $c_M$th column of the base weight mask in a binary mask or its corresponding sampling feature can be calculated as:

$$r_S \in \left\{ r \in \mathbb{Z} : \left\lfloor \frac{(r_M - 1) H_S}{H_M} + \frac{1}{2} \right\rfloor + 1 \leq r \leq \left\lfloor \frac{r_M H_S}{H_M} + \frac{1}{2} \right\rfloor \right\} \quad (7)$$

$$c_S \in \left\{ c \in \mathbb{Z} : \left\lfloor \frac{(c_M - 1) W_S}{W_M} + \frac{1}{2} \right\rfloor + 1 \leq c \leq \left\lfloor \frac{c_M W_S}{W_M} + \frac{1}{2} \right\rfloor \right\} \quad (8)$$

where $H_M \times W_M$ is the size of the base weight mask and $H_S \times W_S$ is the size of a generated binary mask or its corresponding sampling feature. The reference mask $M$ is selected according to the principle that the impact of the discriminative features can be decreased effectively. In practice, given a positive sampling feature, we select a binary mask owning the same size with the base weight mask and all values in the binary mask set 1. Then the value in each position of the selected binary mask is set to 0 in turn to generates $H_M \times W_M$ binary masks $\widehat{M}$. The operation $f_1(\cdot)$ is defined to achieve binary masks $\widehat{M}_s$ with suitable sizes for sampling features. Compared to $f_2(\cdot)$, $f_1(\cdot)$ just set the values in $\widehat{M}_s$ at specific positions, which correspond to the positions owning 0 values in $\widehat{M}$, to 0 by Equation (7) and (8). The sampling features are diversified by $\widehat{M}_s$ through the dropout operation. These diversified sampling features are passed into the classification network $D$, and we pick up the one with the highest loss. The mask in $\widehat{M}_s$ corresponding to the picked up diversified sampling feature is set as $M$.

In Equation (5), the adversarial learning is integrated into our proposed tracking-by-detection framework based on sampling deep convolutional features. Since the features of the samples are sampling in deep convolutional features with lower resolutions, the positive samples in each frame are highly spatially overlapped, and they may fail to capture rich appearance variations. So adversarial feature generator network $G$ is used to produce masks which represent different types of appearance variation. Though the adversarial learning process, $G$ will gradually produce adversarial sampling features that degrades the classifier most. The classification network $D$ will gradually be trained without overfitting to the discriminative features from individual frames while relying on more robust features over a long temporal span.

### 3.4 Tracking Process

The tracking process of our proposed AFSL-based tracker includes three parts: model initialization, online detection and online model update. The details are as follows:

(1) Model initialization.

The based deep CNN is initialized by the VGG-16 model [17] trained in on the ILSVRC-2012 dataset. The parameters in C1, C2, and C3 of the base deep is fixed. C4 and C5 of the base deep CNN and the classification network is offline pretrained by the multi-domain learning strategy shown in [10].

(2) Online detection.

Given an input frame, we first generate multiple candidate boxes according the position of the target object in the previous frame. Unlike the existing tracking-by-detection approaches, the based deep CNN only running one time and the CNN features corresponding to the candidate boxes are extracted by sampling in the output of the based deep CNN. The adversarial feature generator network is removed and we feed the sampling CNN features of the candidate proposals into the classification network to get the probability scores.

(3) Online model update

Around the given position in the first frame or the estimated position in other frames, we generate multiple candidate boxes and assign them with binary labels according to their intersection-over-union scores with the given or estimated bounding box. Then these labeled training samples are used to jointly train adversarial feature generator network $G$ and the classification network $D$. In one iteration of the training process, we pass the positive sampling feature SBRF1 through $G$ and obtain the predicted base weight mask. Then the operation $f_2(\cdot)$ defined in Section 3.3 is used to generate binary masks with suitable sizes for the sampling feature SBRF1 and SBRF2. The adaptive dropout operations are conducted on SBRF1 and SBRF2 through the binary masks to generated the adversarial positive sampling features AF1 and AF2. We keep the labels unchanged and train $D$ through supervised learning. After training $D$ once, given a positive sampling feature, we used the process described in Section 3.3 to obtain the reference mask $M$ in Equation (5) and update $G$ accordingly.

### 4 Experiments

In this section, the implementation details of our proposed AFSL-based tracker are introduced firstly and then we compare our AFSL-based tracker with state-of-the-art trackers on the benchmark datasets

OTB2013 [15], OTB2015 [14], VOT2015 [19] and VOT2016 [16] for performance evaluation.

**4.1 Implementation**

In Section 3.1, the fixed cropped expansion rate $r_c$ is set to 1.2 and the width or the length $L$ of the target object in the cropped and resized image patch is set to 112. In Section 3.3, the size $H_s^1 \times W_s^1$ of the sampling feature SBRF1 is set to $6 \times 6$ and the size $H_s^2 \times W_s^2$ of the sampling feature SBRF2 is set to $8 \times 8$. The size $H_M \times W_M$ of the base weight mask can be deduced as $3 \times 3$ according to architecture of the adversarial feature generator network $G$ introduced in Section 3.3. During the adversarial learning, the SGD solver is iteratively applied to both the adversarial feature generator network $G$ and the classification network $D$ and 60 iterations are used to initialize both networks. We update both networks every 10 frames using 10 iterations. For training $D$, the learning rate is 0.01, the momentum is 0.9 and the weight decay is 0.0005. For training $G$, the learning rate is 0.00005, the momentum is 0.9 and the weight decay is 0.0005. Each mini-batch consists of 32 positives and 96 hard negatives selected out of 1024 negative examples. Our AFSL-based tracker runs on a PC with an Intel(R) Core(TM) i7-5930K CPU @ 3.50GHz clock, 32GB RAM and a NVIDIA TITAN Xp GPU using MATLAB R2017a and MatConvNet toolbox [20]. The tracking speed of AFSL is around 12.1 FPS.

**4.2 Evaluation on OTB**

OTB-2015 [14] contains 100 fully annotated videos with many tracking challenges including occlusion, deformation, scale variation, fast motion and so on. OTB-2013 [15], which is a subset of OTB-2015, includes 50 fully annotated videos. The tracking performance was measured by conducting a one-pass evaluation (OPE) based on two metrics: precision plot and success plot. The precision plot shows the frame locations rate within a certain threshold distance from ground truth locations. The rate corresponding to the threshold distance of 20 pixels is always taken as the representative precision score. The success plot metric is set to measure the overlap ratio between the predicted bounding boxes and the ground truth.

We compare our method against 10 recent state-of-the-art methods including VITAL (CVPR 2018) [7], ADNet (CVPR 2017) [8], CREST (ICCV 2017) [9], MCPF (CVPR 2017) [29], MDNet (CVPR 2016) [10], SINT (CVPR 2016) [39], HDT (CVPR 2016) [28], SRDCFdecon (CVPR 2016) [40], DeepSRDCF (ICCV 2015 Workshop) [41], CNN-SVM (ICML 2015) [12].

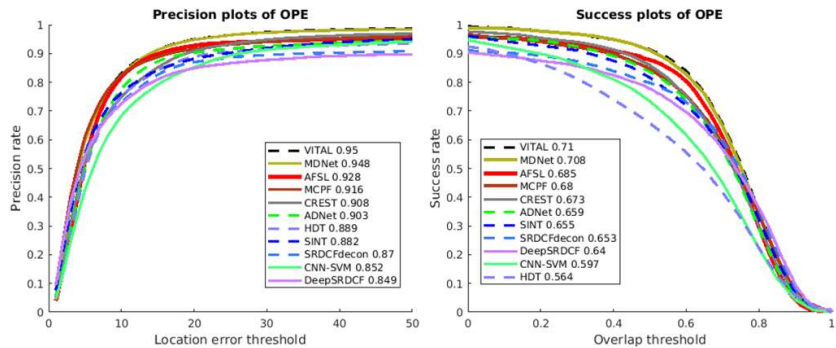

**Figure 3. Precision and success plots on OTB2013.** The values in the legend indicate the representative precisions at 20 pixels for precision plots, and the area-under-curve scores for success plots.

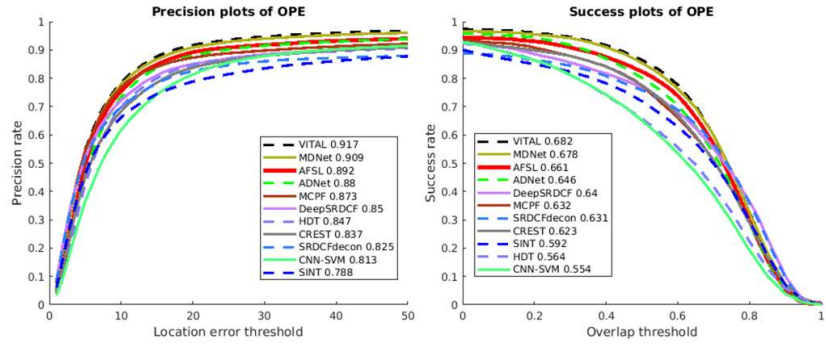

**Figure 4. Precision and success plots on OTB2015.** The values in the legend indicate the representative precisions at 20 pixels for precision plots, and the area-under-curve scores for success plots.

Fig.3 and Fig.4 illustrates the precision and success plots for OTB2013 and OTB2015 respectively. It clearly illustrates that our algorithm, denoted by AFSL, outperforms most of recent state-of-the-art trackers significantly in both measures. The performances of VITAL and MDNet are slighter better than AFSL. The main reason is that VITAL and MDNet draw samples in the raw-image and their deep networks can concentrate on processing each sample and own better advantage for OPE. Though the VITAL and MDNet's precision and success plots of OPE are slightly better than AFSL, the running speed of AFSL is about 8.6 times and 7.6 times faster than VITAL and MDNet respectively. The speeds of MDNet and VITAL is 1.6 and 1.4 frames per second (FPS) respectively. The speed of AFSL is 12.1 FPS. The main reason of AFSL's speed boost is that only one cropped image around the target object is input into the deep CNN and the samples is sampled on its feature maps. In VITAL and MDNet methods, quite a number of samples are drawn in the raw-image and input into the deep CNN, which is time consuming.

In Figure 5, we further analyze the tracking performance under different video attributes (e.g., illumination variation, in-plain rotation, low resolution, out-of-plane rotation, background clutter, motion blur) annotated in the OTB2015 benchmark. We show the OPE on the AUC score under six main video attributes. The results indicate that our AFSL tracker is effective in handling illumination variation, in-plain rotation, low resolution, out-of-plane rotation. It is mainly because the adversarial learning integrated in AFSL can enrich the target appearances in the feature space and augment the positive samples by predicting discriminative features according to different inputs.

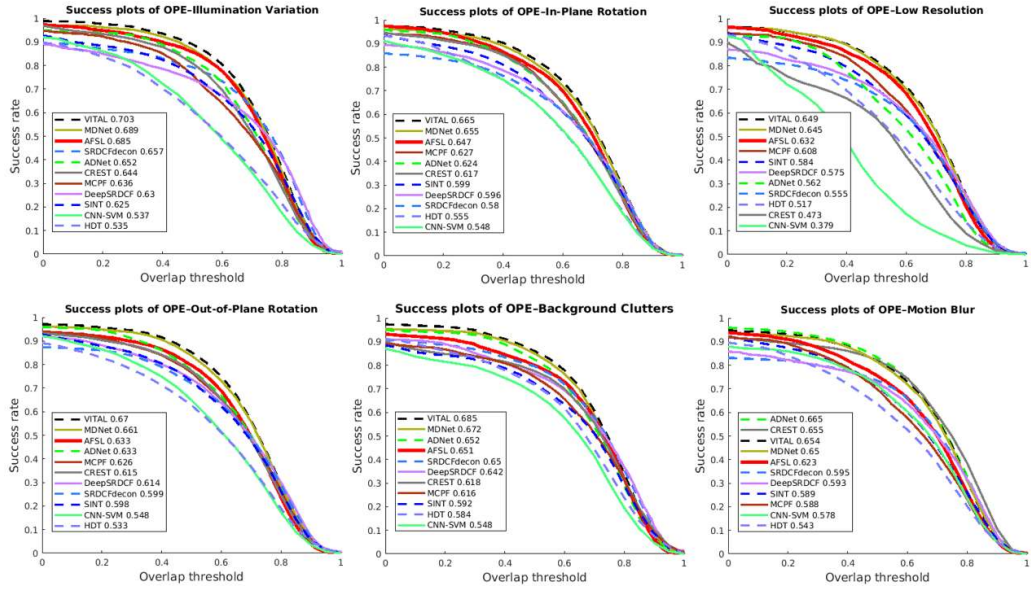

**Figure 5**: The success plots for six challenge attributes of OTB2015: illumination variation, in-plain rotation, low resolution, out-of-plane rotation, background clutter, motion blur.

**4.3 Evaluation on VOT2015**

The VOT2015 [29] benchmark contains results of 63 state-of-the-art trackers evaluated on 60 challenging sequences. The VOT performance evaluation methodology [42] detects a tracking failure and re-initializes the tracker when the tracker drifts off the target. The VOT mainly adopts three metrics for performance evaluation: average number of failures during tracking (robustness), average overlap during the periods of successful tracking (accuracy) and expected average overlap (EAO) on short-term sequences.

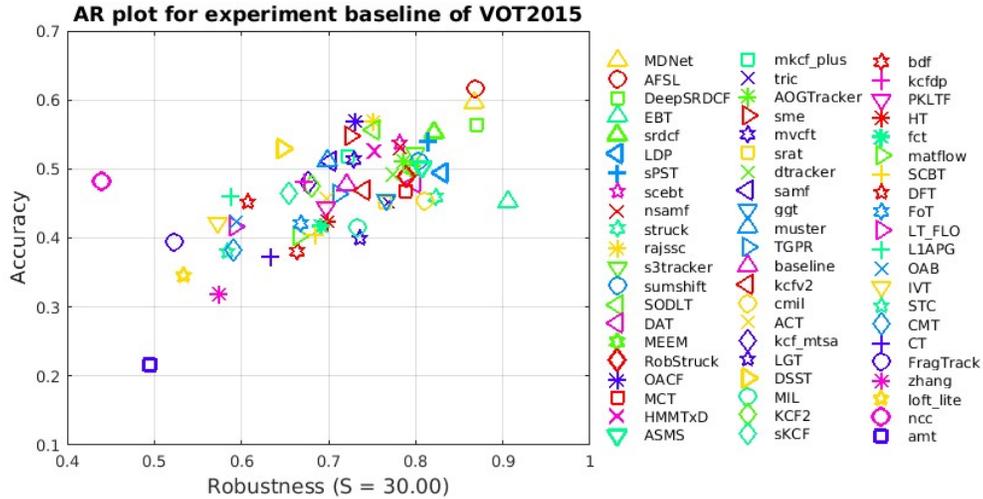

**Figure 6**: The AR plot for the experiment baseline of VOT2015. The better trackers are located at the upper-right corner.

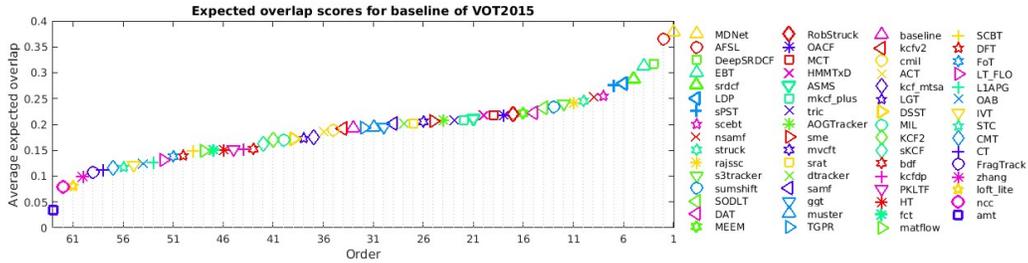

**Figure 7**: EAO ranking with trackers in VOT2015. The better trackers are located at the right. The better trackers are located at the right.

In Figure 6, we show the AR plot for the experiment baseline of VOT2015 and our AFSL achieve the best accuracy. Our method obtains an accuracy of 0.62, outperforms MDNet [10] (first place in EAO criterion) and DeepSRDCF [41] (third place in EAO criterion) by 2% and 6%. Figure7 illustrates that our AFSL can rank second in the 63 trackers of VOT2015 benchmark according to EAO criterion. Though the EAO of MDNet outperforms our method by 1.2%, the running speed of AFSL is about 7.6 times faster than MDNet.

In order to further verify the better performance of our AFSL, we compare our method with other recent state-of-the-art methods including FlowTrack (CVPR 2018) [43], BranchOut (CVPR 2017) [44], CSR-DCF (CVPR 2017) [45] and CCOT (ECCV 2016) [46]. As shown in Table1, our method achieves best performance in EAO and accuracy among these four state-of-the-art trackers.

**Table 1**: State-of-the-art results on VOT2015 dataset. EAO: expected average overlap; A: accuracy; R: robustness (average number of failures).

|  |  | EAO | A | R |
|---|---|---|---|---|
|  | AFSL | 0.366 | 0.62 | 0.98 |
| Other recent trackers | FlowTrack [43] | 0.341 | 0.57 | 0.95 |
|  | BranchOut [44] | 0.338 | 0.59 | 0.71 |
|  | CSR-DCF [45] | 0.338 | 0.51 | 0.85 |
|  | CCOT [46] | 0.331 | 0.52 | 0.85 |
| Top trackers in VOT2015 | MDNet [10] | 0.378 | 0.60 | 0.77 |
|  | DeepSRDCF [41] | 0.318 | 0.56 | 1.00 |
|  | EBT [47] | 0.313 | 0.45 | 0.81 |
|  | srdcf [48] | 0.288 | 0.55 | 1.18 |
|  | LDP [49] | 0.279 | 0.49 | 1.30 |
|  | sPST [50] | 0.277 | 0.54 | 1.42 |

**4.4 Evaluation on VOT2016**

The datasets in VOT2016 [16] are the same as VOT2015, but the ground truth bounding boxes in the VOT2015 dataset have been re-annotated and the bounding box of each frame is automatically generated from the segmentation mask which has been manually per-pixel segmented. The VOT2016 benchmark contains results of 70 state-of-the-art trackers evaluated on the 60 re-annotated challenging sequences. VOT2016 also adopts EAO, accuracy and robustness for evaluations.

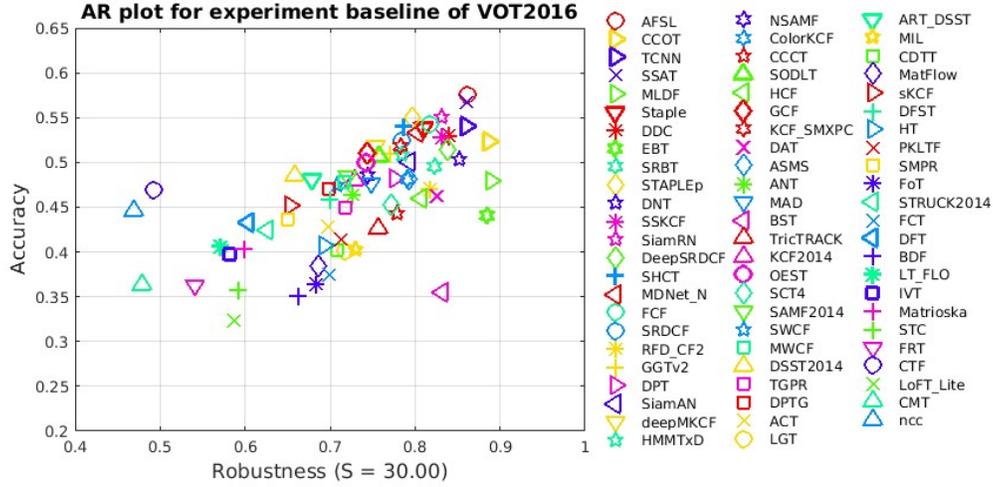

**Figure 8**: The AR plot for the experiment baseline of VOT2016. The better trackers are located at the upper-right corner.

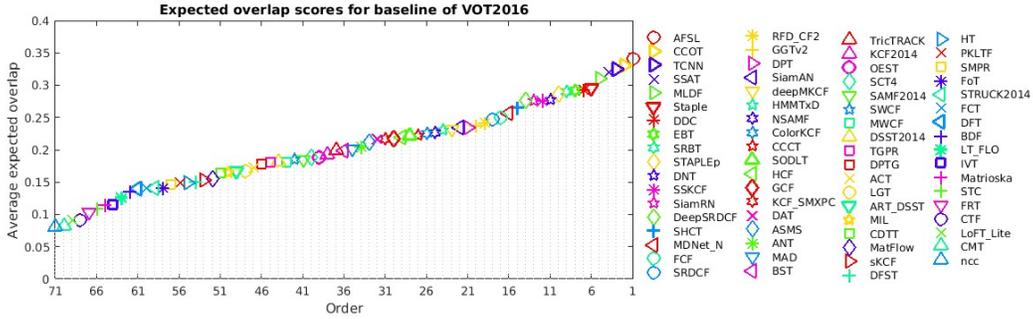

**Figure 9**: EAO ranking with trackers in VOT2016. The better trackers are located at the right.

In Figure 8, we show the AR plot for the experiment baseline of VOT2016 and our AFSL achieve the best accuracy. Our method obtains an accuracy of 0.58, outperforms CCOT [46] (second place in EAO criterion) and TCNN [51] (third place in EAO criterion) by 6% and 4%. Figure9 illustrates that our AFSL can rank first in the 70 trackers of VOT2016 benchmark according to EAO criterion. It is worth noting that our method can operate at 12.1 FPS, which is 40.3 times faster than CCOT (0.3 FPS).

We also compare our method with more recent state-of-the-art methods including VITAL (CVPR 2018) [7], FlowTrack (CVPR 2018) [43], CREST (ICCV 2017) [9], TSN (ICCV 2017) [52]. As shown in Table2, our method achieves best performance in EAO and accuracy among these four state-of-the-art trackers.

**Table 2**: State-of-the-art results on VOT2016 dataset. EAO: expected average overlap; A: accuracy; R: robustness (average number of failures).

|  |  | EAO | A | R |
|---|---|---|---|---|
|  | AFSL | 0.342 | 0.58 | 1.01 |
| Other recent trackers | VITAL [7] | 0.323 | 0.55 | 0.98 |
|  | FlowTrack [43] | 0.334 | 0.58 | 0.86 |
|  | CREST [9] | 0.283 | 0.51 | 1.08 |
|  | TSN [52] | 0.336 | 0.58 | 0.95 |
| Top trackers in VOT2016 | CCOT [46] | 0.331 | 0.52 | 0.85 |

| | | | |
|---|---|---|---|
| TCNN [51] | 0.325 | 0.54 | 0.96 |
| SSAT [16] | 0.321 | 0.57 | 1.04 |
| MLDF [16] | 0.311 | 0.48 | <span style="color:red">0.83</span> |
| Staple [53] | 0.295 | 0.54 | 1.35 |
| DDC [16] | 0.293 | 0.53 | 1.23 |

**4.5 Qualitative Evaluation**

To visualize the superiority of our proposed AFSL-based tracker, we show some examples' results of our method compared to recent trackers on challenging sample videos. Figure 10 shows some results of the top performing trackers: VITAL [7], MDNet [10], ADNet [8], MCPF [29], CREST [9] and our proposed AFSL-based tracker on five representative challenging sequences.

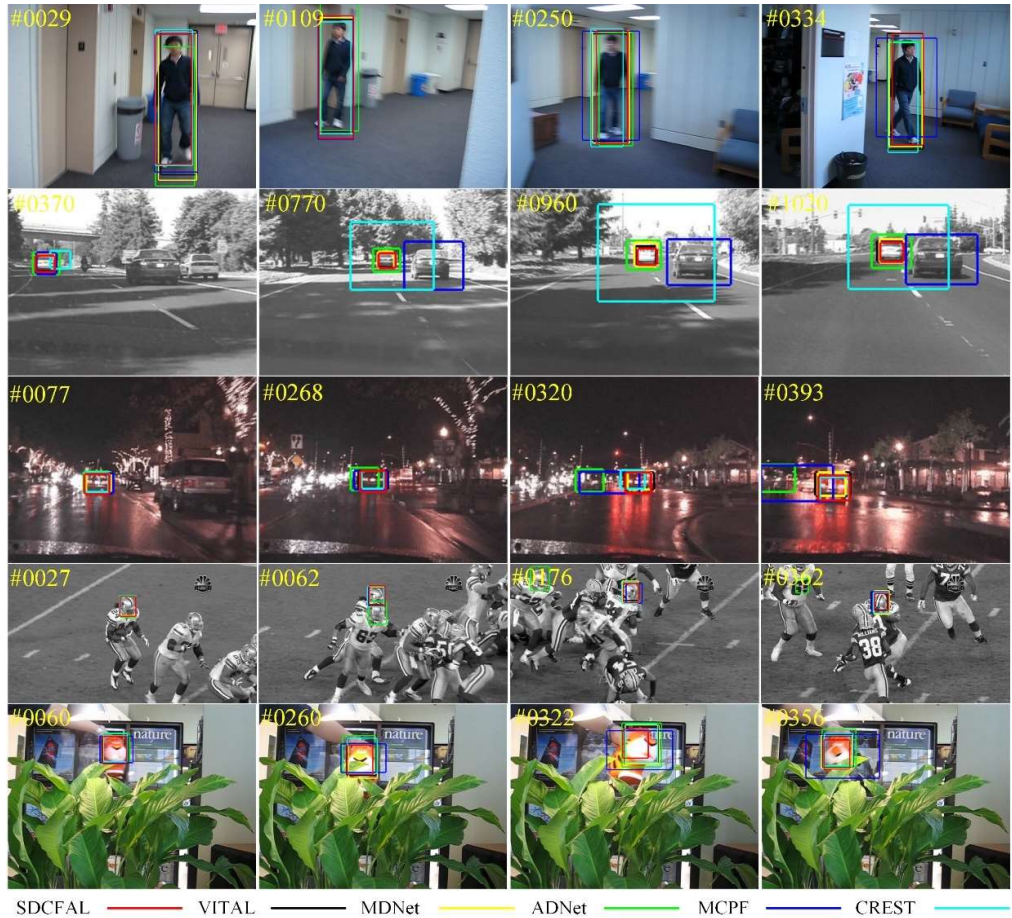

**Figure 10**: Tracking results comparison of our approach with four state-of-the-art trackers in the challenging scenarios. The video sequences are BlurBody, Car1, CarDark, Football and Tiger2 respectively from top to down.

As show in Figure.10, the target in sequence Car1 undergoes illumination variation, scale variation, background clutters and low resolution. CREST lose the target from #370, MCPF lose the target from #700 and ADNet can't fit the scale change. In contrast, the proposed AFSL-based tracker achieves successful tracking in this sequence because the adversarial learning integrated in our tracking framework can enrich the target appearances described by sampling deep convolutional features and augment the positive samples. The challenges in sequence BlurBody include fast relative motion, in-plane rotation and motion blur, and the proposed method can handle these challenges while MCPF can't achieve high overlap with the ground truth from #250. CarDark is a sequence with challenging illumination variation and background clutters, and ADNet and MCPF can't track the target successfully from #320. There are challenges such as occlusion, in-plane rotation, out-of-plane rotation, background clutters in the sequence Football, and ADNet loses the target from #62. The challenges in sequence Tiger2 including illumination variation, occlusion, deformation, motion blur, fast motion, in-plane rotation, out-of-plane rotation and out-of-view, and MCPF can't achieve high overlap with the ground truth. Our proposed AFSL-based tracker, VITAL and MDNet can track targets in these challenging videos successfully and accurately. What's more, compared with VITAL and MDNet, our AFSL-based tracker achieves faster speed.

## 5. Conclusions

In this paper, we propose an efficient visual tracking based on adversarial feature sampling learning. The proposed method changes the strategy of obtaining samples around the target object in the traditional tracking-by-detection framework. It draws samples on the feature maps of the deep convolution network rather than on the draw-image, which accelerates tracking-by-detection trackers based on raw-image samples effectively. Furthermore, a generative adversarial network is integrated into our network framework to augment positive samples and improve the tracking performance. Extensive experiments on benchmark datasets demonstrate that the proposed method achieves a comparable performance to state-of-the-art trackers and accelerates traditional deep-learning-based tracking-by-detection trackers effectively.

**Acknowledgment**

This work was supported in part by the Hong Kong Scholars Program and in part by the National Natural Science Foundation of China under Grant 61703398. We would like to thank NVIDIA for providing the GPU card.